\documentclass{article}

\usepackage{microtype}
\usepackage{graphicx}
\usepackage{subcaption}
\usepackage{booktabs} 

\usepackage{hyperref}

\usepackage[preprint]{icml2026}

\usepackage{amsmath}
\usepackage{amssymb}
\usepackage{mathtools}
\usepackage{amsthm}

\usepackage[capitalize,noabbrev]{cleveref}

\theoremstyle{plain}

\theoremstyle{definition}

\theoremstyle{remark}

\usepackage[textsize=tiny]{todonotes}

\icmltitlerunning{NIVA: A Multimodal Foundation Model for Actionable Earth System Intelligence}

\begin{document}
\newcommand{\niva}{\textsc{NIVA }}
\newcommand{\AwardNumber}{NSF SBIR Phase I Award \#\,2507255}
\twocolumn[
  \icmltitle{NIVA: A Multimodal Foundation Model for \\Actionable Earth System Intelligence}


  \icmlsetsymbol{equal}{*}

  \begin{icmlauthorlist}
    \icmlauthor{Anisha Pal}{yyy,zzz}
    \icmlauthor{Aodhan Sweeney }{equal,xxx}
    \icmlauthor{Kyle Heyblom}{equal,yyy,zzz}
    \icmlauthor{Kalai Ramea}{yyy,zzz}
  \end{icmlauthorlist}

  \icmlaffiliation{yyy}{Independent Researcher}
  \icmlaffiliation{xxx}{Planette AI, USA}
  \icmlaffiliation{zzz}{Work done while at Planette AI}

  \icmlcorrespondingauthor{Anisha Pal}{apal72@gatech.edu}
  \icmlcorrespondingauthor{Aodhan Sweeney}{aodhan.sweeney@planette.ai}

  \icmlkeywords{Machine Learning, ICML}

  \vskip 0.3in
]



\printAffiliationsAndNotice{\icmlEqualContribution}

\begin{abstract}
\label{sec-abstract}
Recent advances in AI-driven weather and climate modeling have improved forecast skill while reducing computational cost. However, existing data-driven approaches are limited in their ability to model coupled Earth system dynamics, which is required for extending predictability beyond the $\approx$2-week horizon. To address this, we introduce \niva, a multimodal foundation model designed to learn unified representations across Earth system components. While the full framework targets atmosphere, ocean, ice, and land interactions, we focus here on a two-modality setting (ocean and atmosphere) as a controlled proof of concept to evaluate whether foundation models can learn coupled dynamics. Trained on large-scale Earth system simulations, \niva learns physically meaningful cross-modal structure, providing a foundation for subseasonal-to-seasonal prediction. As initial validation, we show that NIVA captures key modes of climate variability through accurate prediction of major climate indices.

\end{abstract}

\section{Introduction}
\label{sec-intro}

\begin{figure*}[t]
    \centering
    \includegraphics[width=\textwidth]{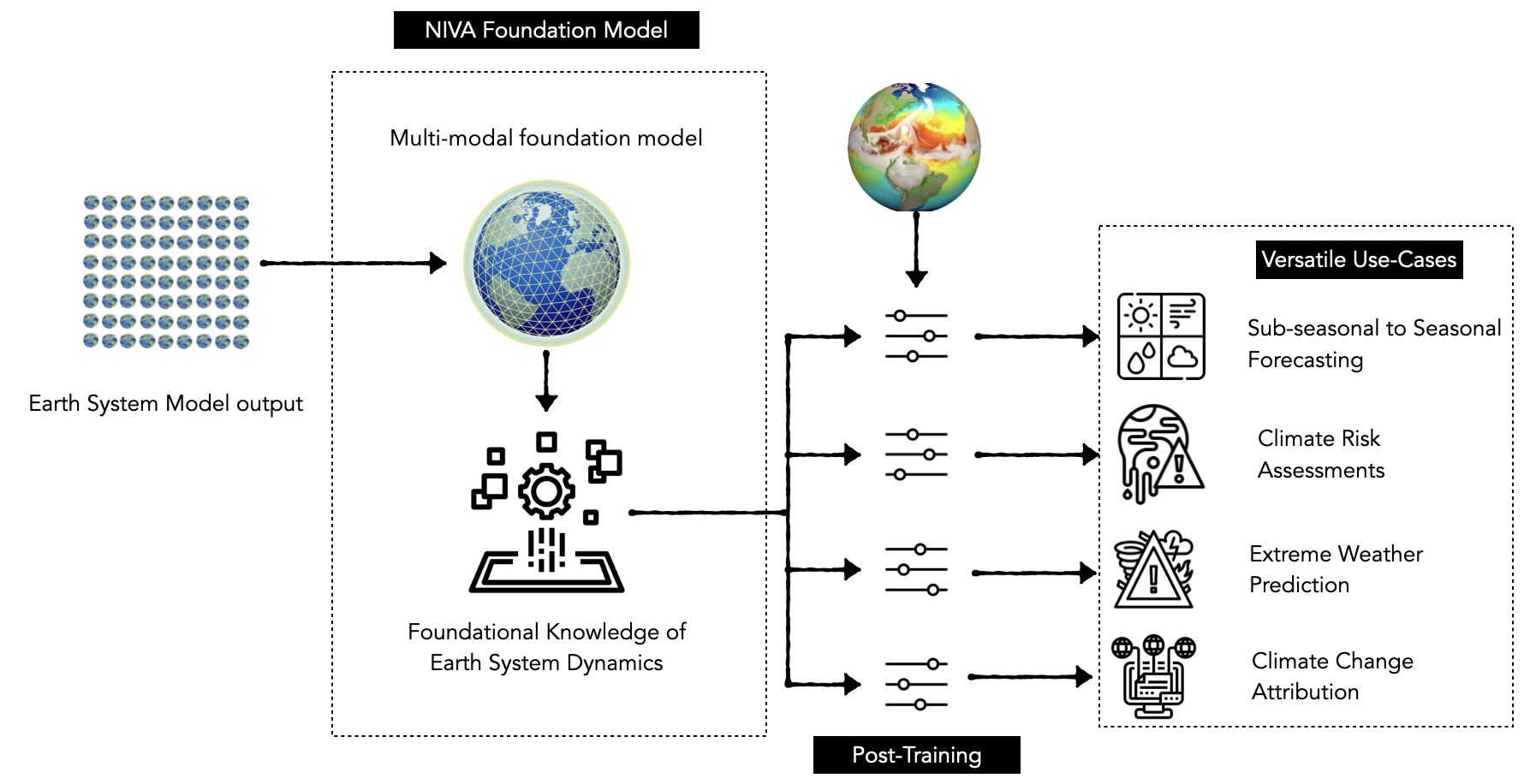}
\caption{\textbf{Schematic of the \niva end-to-end pipeline.} The framework uses ESM output to pretrain the foundation model that learns coupled Earth system representations, which can be transferred to a range of downstream tasks.}
    \label{fig:schematic}
\end{figure*}

Foundation models have emerged as a powerful paradigm for learning general-purpose representations from large-scale data, enabling flexible adaptation across a wide range of downstream tasks. In domains such as natural language processing and computer vision, these models have shifted the focus from task-specific solutions to unified architectures that capture underlying structure in complex systems~\citep{radford2018improving, devlin2019bertpretrainingdeepbidirectional, oquab2024dinov2learningrobustvisual, kirillov2023segment, beyer2024paligemmaversatile3bvlm, cheng2024yoloworldrealtimeopenvocabularyobject, radford2021learningtransferablevisualmodels, touvron2023llamaopenefficientfoundation}. A growing line of work seeks to develop foundation models in the context of Earth system science, where the high-dimensional, nonlinear dynamics of the system, combined with increasing availability of long-term datasets, motivate data-driven approaches to learn these processes directly from observations.

Early Earth system foundation models, including ClimaX~\citep{nguyen2023climaxfoundationmodelweather}, AtmoRep~\citep{lessig2023atmorepstochasticmodelatmosphere}, Aurora~\citep{bodnar2024foundationmodelearth}, and Prithvi WxC~\citep{schmude2024prithviwxcfoundationmodel}, demonstrate the promise of learning task-agnostic representations from large-scale atmospheric data. These models support a range of downstream applications, including forecasting, downscaling, and event counterfactuals, often rivaling traditional numerical approaches in short-range prediction tasks. However, despite these advances, existing models remain limited in scope, as they primarily focus on atmospheric variables and do not fully capture the coupled dynamics of the broader Earth system, such as the ocean, sea ice, and land surface. As a result, restricting their predictive skill to shorter timescales, as atmospheric predictability beyond $\approx 2$ weeks depends on memory stored in these slow-evolving components ~\citep{oort1992physics}.

Furthermore, existing models are largely trained on observational reanalysis datasets, which are dominated by atmospheric observations and data assimilation systems. This leads to an atmosphere-centric view of the Earth system, in which slowly evolving components such as the ocean are under observed and weakly constrained, especially at the timescales relevant to their variability. Moreover, training on high temporal resolution data (e.g., hourly) biases these models toward fast atmospheric processes and short-term variability, limiting their ability to capture slower changes and interactions between different parts of the Earth system that evolve over weekly to seasonal timescales.

This limitation is critical because predictability at subseasonal-to-seasonal (S2S) timescales depends on interactions between the atmosphere and slower components such as the ocean, sea ice, and land surface~\citep{oort1992physics}. As a result, by emphasizing high-frequency atmospheric dynamics while underrepresenting lower-frequency variability and coupling, current foundation models struggle to address key challenges such as S2S forecasting and large-scale climate variability.

To address this gap in existing work, we introduce NIVA, a multimodal foundation model designed to learn coupled Earth system dynamics. Drawing on methodologies from vision–language architectures~\citep{cheng2024yoloworldrealtimeopenvocabularyobject, he2021maskedautoencodersscalablevision, singh2022flavafoundationallanguagevision, jia2021scalingvisualvisionlanguagerepresentation}, \niva learns shared high-dimensional representations across heterogeneous components, enabling it to model interactions between the atmosphere, ocean, sea ice, and land surface. 

\par
To further ground our approach, we frame this work around the following core questions:
\begin{itemize}
    \item Can multimodal foundation models learn unified representations that capture the coupled dynamics across Earth system components?
    \item To what extent can joint ocean--atmosphere representation learning recover physically meaningful modes of climate variability?
    \item Are numerical simulations of the Earth system an effective pretraining data source for learning coupled dynamics that transfer to the observational domain?
\end{itemize}
Answering these questions provides insight into how data-driven AI models can capture coupled Earth system dynamics, which is critical to extend predictability beyond the current atmospheric predictability limit of $\approx 2$ weeks. Our overarching goal is to develop a foundation model that learns a physically meaningful and generalizable latent space encoding interactions across Earth system components. 

In this work, we focus on the ocean–atmosphere subsystem, which is central to S2S predictability due to the ocean’s longer intrinsic memory and its role in modulating large-scale atmospheric variability. We show that \niva captures these interactions by accurately predicting major modes of climate variability, providing evidence that the learned latent space encodes physically meaningful coupled dynamics.

Our results suggest that multimodal representation learning can recover key drivers of large-scale climate variability directly from data, offering a pathway toward modeling coupled Earth system processes in a data-driven framework. The learned latent space is designed to support transfer to downstream tasks such as climate risk assessment, extreme event prediction, and climate attribution; we leave a systematic evaluation of these capabilities to future work. In addition, we also develop \niva as a modular and extensible framework, enabling the scalable integration of additional Earth system components in future extensions.



\section{Related Works}
\label{sec-background}
\subsection{Numerical Earth System Models}

Numerical Earth System Models (ESMs) are the primary tool for simulating and predicting the behavior of Earth’s climate and weather system. They represent the coupled dynamics of the atmosphere, ocean, land surface, and cryosphere by numerically integrating the governing physical equations, including fluid dynamics, thermodynamics, and radiative transfer, on discrete spatial grids \citep{flato2011}.

Numerical ESMs underpin a wide range of scientific and operational applications. In numerical weather prediction (NWP), they are used to generate forecasts from hours to weeks ahead by assimilating observational data into initial conditions and evolving the system forward in time \citep{bauer2015}. At longer timescales, ESMs are central to climate risk assessment and projection, where they are used to simulate future scenarios under different emissions pathways \citep{oneill2016}. They also play a key role in climate change attribution, enabling controlled experiments that isolate the effects of specific drivers like greenhouse gas emissions 
or aerosols \citep{stott2010}. 


Despite their central role, numerical ESMs have several well-known limitations. They are computationally expensive, often requiring high-performance computing infrastructure to run \citep{balaji2017,acosta2024}. Their development and maintenance involve complex codebases, many of which have evolved over decades and are implemented in legacy languages such as Fortran \citep{bauer2021}. Furthermore, key processes that occur below the grid scale, such as convection and cloud microphysics, must be represented through empirical parameterizations, which limits fidelity \citep{schneider2017}. Building, tuning, and interpreting these models requires substantial domain expertise across numerous disciplines, creating a high barrier to entry and slows iteration \citep{balaji2017}.

\subsection{Multimodal Representation Learning}
\label{subsec-mrl}
Multimodal representation learning integrates information from heterogeneous data sources to derive contextually rich semantic representations that capture relationships across modalities~\citep{baltrušaitis2017multimodalmachinelearningsurvey}. By learning shared latent spaces, these models support cross-modal tasks such as retrieval, alignment, and reasoning~\citep{10.5555/2886521.2886647,lu2019vilbertpretrainingtaskagnosticvisiolinguistic, radford2021learningtransferablevisualmodels}. Recent advances, particularly contrastive approaches such as CLIP~\citep{radford2021learningtransferablevisualmodels} and ALIGN~\citep{jia2021scalingvisualvisionlanguagerepresentation}, have demonstrated that large-scale pretraining on paired data yields highly generalizable representations by pulling semantically related samples together while pushing unrelated ones apart. This formulation is especially well-suited to capturing asymmetric, probabilistic relationships between modalities. Despite the broad adoption of contrastive learning across diverse domains and tasks~\cite{oord2019representationlearningcontrastivepredictive,elizalde2022claplearningaudioconcepts,  yue2022ts2vecuniversalrepresentationtime,wu2026autoaugmentationcontrastivelearningwearablebased}, the predominant focus remains on vision-language modalities. In this work, we extend multimodal representation learning to the Earth system domain to model the coupled dynamics between oceanic and atmospheric states by learning a shared latent space that encodes a generalizable and 
physically consistent representation of the Earth system.


\subsection{Foundation Models for Weather and Climate}
As discussed in Sec.~\ref{sec-intro}, while foundation models for weather \& climate learn generalized representations supporting a range of downstream tasks, they remain limited to shorter timescales ~\citep{nguyen2023climaxfoundationmodelweather, lessig2023atmorepstochasticmodelatmosphere, bodnar2024foundationmodelearth, schmude2024prithviwxcfoundationmodel}. 
This is largely because they model the Earth system as an atmosphere-only problem, ignoring slower components like the ocean, sea ice, and land ice that govern long-term memory~\citep{oort1992physics}. 
\par 

Although Aurora~\cite{bodnar2024foundationmodelearth} and Prithvi WxC~\citep{schmude2024prithviwxcfoundationmodel} try to address this gap by training on heterogeneous variables and datasets spanning multiple components of the Earth system, they both rely on a single-stream formulation that effectively treats the system as a unified input space. This design does not account for the asymmetric dependencies or bidirectional couplings between the different subsystems, which are central to Earth system dynamics. Furthermore, their reliance on short temporal contexts limits their ability to represent long-term memory and capture meaningful representations from the slow-moving components of the Earth system. Consequently, while they improve scalability and cross-task generalization, they do not address the core challenge of learning coupled Earth system dynamics. 

\niva addresses these limitations by reformulating Earth system modeling as a multimodal representation learning problem, treating different components as distinct modalities with separate encoders and learning a shared latent space through contrastive alignment, thereby explicitly capturing cross-component dependencies and enabling representations that reflect the coupled nature of the Earth system.


\section{\niva}
\label{sec-methods}
\niva follows a two-stage pipeline consisting of pretraining and post-training (fine-tuning). During pretraining, NIVA uses a two-modality joint learning framework for the ocean–atmosphere system, analogous to image–text representation learning in multimodal models (see Sec.~\ref {subsec-mrl}). This analogy is motivated by the statistical asymmetry between modalities: a single text reference (e.g., “dog”) can correspond to many valid images, and similarly, a slowly evolving ocean state can correspond to multiple dynamically consistent atmospheric realizations due to higher atmospheric variability. NIVA leverages this structure by learning a shared latent space that models conditional relationships and cross-modal constraints between the two systems.

\subsection{Data}
\label{subsec-data}
As discussed in Sec.\ref{sec-intro}, existing foundation models for weather and climate are predominantly trained on high temporal resolution data, which biases them toward fast atmospheric processes and limits their ability to capture lower-frequency modes of variability. This, in turn, restricts their capacity to model the coupled interactions across Earth system components that govern long-term behavior. NIVA addresses this limitation by operating at lower temporal resolutions, enabling the model to better represent slowly evolving subsystems and their interactions. However, foundation model pretraining is inherently data-intensive. Observational datasets, while physically grounded, are limited in temporal extent (typically $<50$ years) and therefore underrepresent slower components such as the ocean, sea ice, land ice, and land surface. To overcome this limitation, we leverage large-scale simulations from the Community Earth System Model version 2 Large Ensemble (CESM2-LE) for pretraining, which provide long, consistent, and fully coupled representations of the Earth system. In contrast, post-training requires significantly less data and benefits from real-world grounding. Therefore, we use the observation-informed dataset ERA5 for downstream tasks and evaluation. This section describes the datasets used for both pretraining and post-training.
\subsubsection{CESM2 Large Ensemble}
CESM2 is a state-of-the-art coupled Earth System Model that simulates interactions among the atmosphere, ocean, land, sea ice, and biosphere \citep{danabasoglu2020}. The CESM2-LE comprises 100 separate simulations (ensemble members) spanning $1850–2100$, combining a historical experiment ($1850–2014$) and an SSP3-7.0 future emissions scenario ($2015–2100$) \citep{rodgers2021}.  Each ensemble member simulation follows an almost identical external forcing pathway but starts from slightly different initial atmospheric and ocean states, producing 100 plausible but distinct sequences of weather and climate\citep{https://doi.org/10.1029/2021GL097420}. The training dataset includes atmospheric, oceanic, and invariant variables selected to capture the major modes of coupled Earth system variability (see Appendix \ref{appendix-variables} for details). All variables are represented on a common $1^\circ$ x $1^\circ$ longitude–latitude grid and follow consistent spatial and temporal conventions. We restrict training to ensemble members spanning $1920–2100$ to align the forcing distributions with the evaluation period. Derived variables are included to augment the representation of dynamical and thermodynamical processes. All data was preprocessed into standardized, aggregated, and detrended datasets suitable for training the NIVA foundation model (see Appendix \ref{appendix-preprocessing}).

\subsubsection{ERA5}
The ERA5 reanalysis dataset \citep{hersbach2020} is produced by the European Centre for Medium-Range Weather Forecasts. ERA5 provides a globally complete, observation-constrained estimate of the atmospheric state by assimilating a wide range of satellite and in-situ measurements into a numerical weather prediction system. It offers high temporal resolution and consistent global coverage over multiple decades, making it well suited for evaluating model performance under real-world conditions and for bridging the gap between model-simulated and observed Earth system variability. ERA5 data used for post-training are all represented on a common $0.25^\circ$ x $0.25^\circ$ longitude–latitude grid and follow consistent spatial and temporal conventions. Data from years $1980$ to $2025$ were used. For preprocessing steps, refer to Appendix \ref{appendix-preprocessing}.

\subsection{Pretraining}
\begin{figure*}[t]
    \centering
    \includegraphics[width=\textwidth]{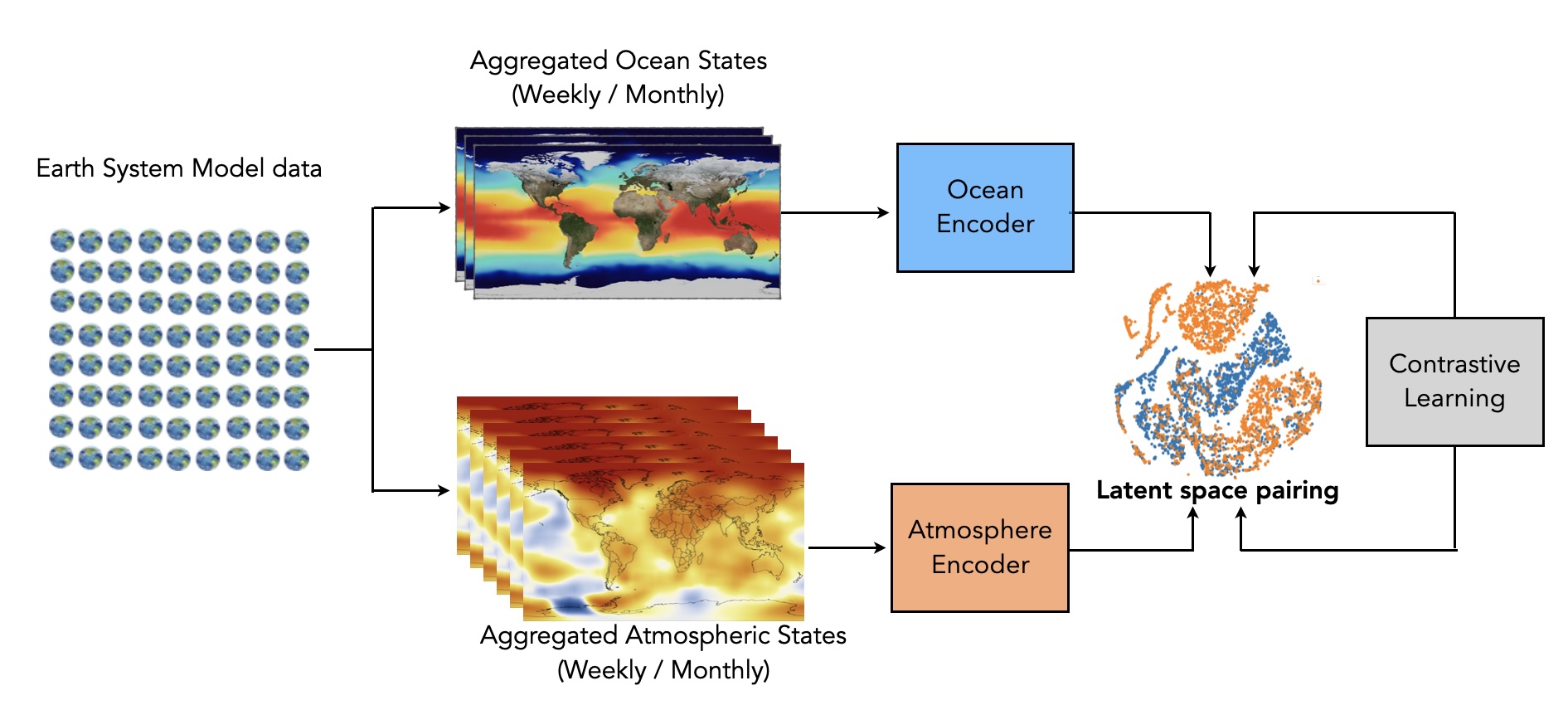}
    \caption{\textbf{Pretraining approach.} Aggregated oceanic and atmospheric states are processed by separate encoders and jointly optimized with a contrastive objective to learn a shared latent representation of coupled dynamics.}
    \label{fig:arch}
\end{figure*}
\label{subsec-pretraining}
Our pretraining framework follows a contrastive learning objective with separate encoders for oceanic and atmospheric states. The encoders learn modality-specific representations while aligning paired samples in a shared latent space to capture cross-modal relationships.

\subsubsection{Pretraining Objective}

As illustrated in Fig.~\ref{fig:arch}, both the oceanic and atmospheric encoders take as input tensors of size $(C, 128, 256)$ at each timestep, where $C$ denotes the number of channels(input variables). A timestep is defined as a temporal aggregation window (e.g., weekly or monthly), such that each datapoint represents a spatial snapshot of aggregated oceanic or atmospheric variables over that interval. For our initial formulation, oceanic and atmospheric data are defined at the same temporal resolution (e.g., both monthly or both weekly), positive pairs are constructed by aligning ocean and atmosphere states from the same timestep. Negative pairs are sampled from non-overlapping timesteps, enforcing temporal inconsistency (for additional details on other formulations, see Sec.~\ref{appendix-pretraining}).  
\par
Following encoding, the model is optimized using a contrastive objective that encourages aligned ocean–atmosphere pairs to be close in the latent space while pushing misaligned pairs apart. Positive pairs are defined based on temporal co-occurrence within a shared aggregation window, while negative pairs are drawn from temporally disjoint windows. This formulation is agnostic to the specific temporal resolution and supports both one-to-one and one-to-many alignments. The similarity metric (cosine similarity) and corresponding loss formulation for this contrastive objective is detailed in Sec.~\ref{subsec-loss}. Through this self-supervised objective, the model learns a shared latent space that captures the conditional structure and cross-modal dependencies governing ocean–atmosphere interactions.
\par
In the current implementation of \niva, we restrict training to a monthly temporal resolution with a one-to-one alignment between oceanic and atmospheric states. This design choice simplifies the learning problem while preserving the dominant coupled variability at longer timescales. It serves as a foundation for future extensions to multi-resolution and one-to-many training regimes(see Sec.~\ref{appendix-pretraining}).

\subsubsection{Ocean \& Atmospheric Encoder}

We evaluated several encoder architectures for both modalities, including Vision Transformers (ViT) \citep{Dosovitskiy2021_ViT}, Swin Transformer V2\citep{Liu2022_SwinV2}, and Spherical Fourier Neural Operators (SFNO) \citep{Bonev2023_SFNO}. For the final model, we selected SFNO as the encoder for both oceanic and atmospheric modalities, as it achieved the strongest representational performance across variables. This result aligns with SFNO’s capacity to capture global spatial dependencies through its spectral operator formulation. Additional details on the alternative encoder configurations are provided in Sec.~\ref{appendix-encoder}.


\par
While atmospheric variables are continuously defined over the global grid, ocean variables are restricted to ocean-covered regions, introducing spatial sparsity that can lead to training instability in the ocean encoder. We address the sparsity of ocean observations in the SFNO-based ocean encoder by incorporating land-sea fraction mask as an additional input channel, allowing the model to condition its representations on domain geometry. Furthermore, to avoid optimization instability arising from sharp land–ocean discontinuities in the ocean variables, we interpolate them over land regions to produce continuous inputs. Although this introduces artificial values, the explicit mask enables the model to distinguish valid ocean regions from interpolated areas. Finally, to ensure that learned representations are restricted to physically meaningful regions, we apply mask pooling at the output of the final SFNO layer. Let $\mathbf{H} \in \mathbb{R}^{B \times C \times H \times W}$ denote the encoded output feature tensor and let $M \in \{0,1\}^{H \times W}$ be the corresponding land--sea mask, shared across channels. Here, $B$ denotes the batch size, $C$ the number of feature channels, and $H \times W$ the spatial resolution of the feature map.The pooled representation is computed as:
\[
\mathbf{z}_b =
\frac{\sum_{i=1}^{H} \sum_{j=1}^{W} M_{ij} \, \mathbf{H}_b(:, i, j)}
{\sum_{i=1}^{H} \sum_{j=1}^{W} M_{ij} + \epsilon},
\quad \mathbf{z}_b \in \mathbb{R}^{C},
\]
where $\mathbf{H}_b(:, i, j) \in \mathbb{R}^{C}$ denotes the feature vector at spatial location $(i,j)$ for sample $b$. This produces a pooled representation invariant to the number of valid ocean points. The resulting vector is then passed through a linear projection to obtain the final $d$-dimensional latent representation.

\subsubsection{Loss}
\begin{figure*}[t]
    \centering
    
    \begin{subfigure}[t]{0.43\textwidth}
        \centering
        \includegraphics[width=\textwidth]{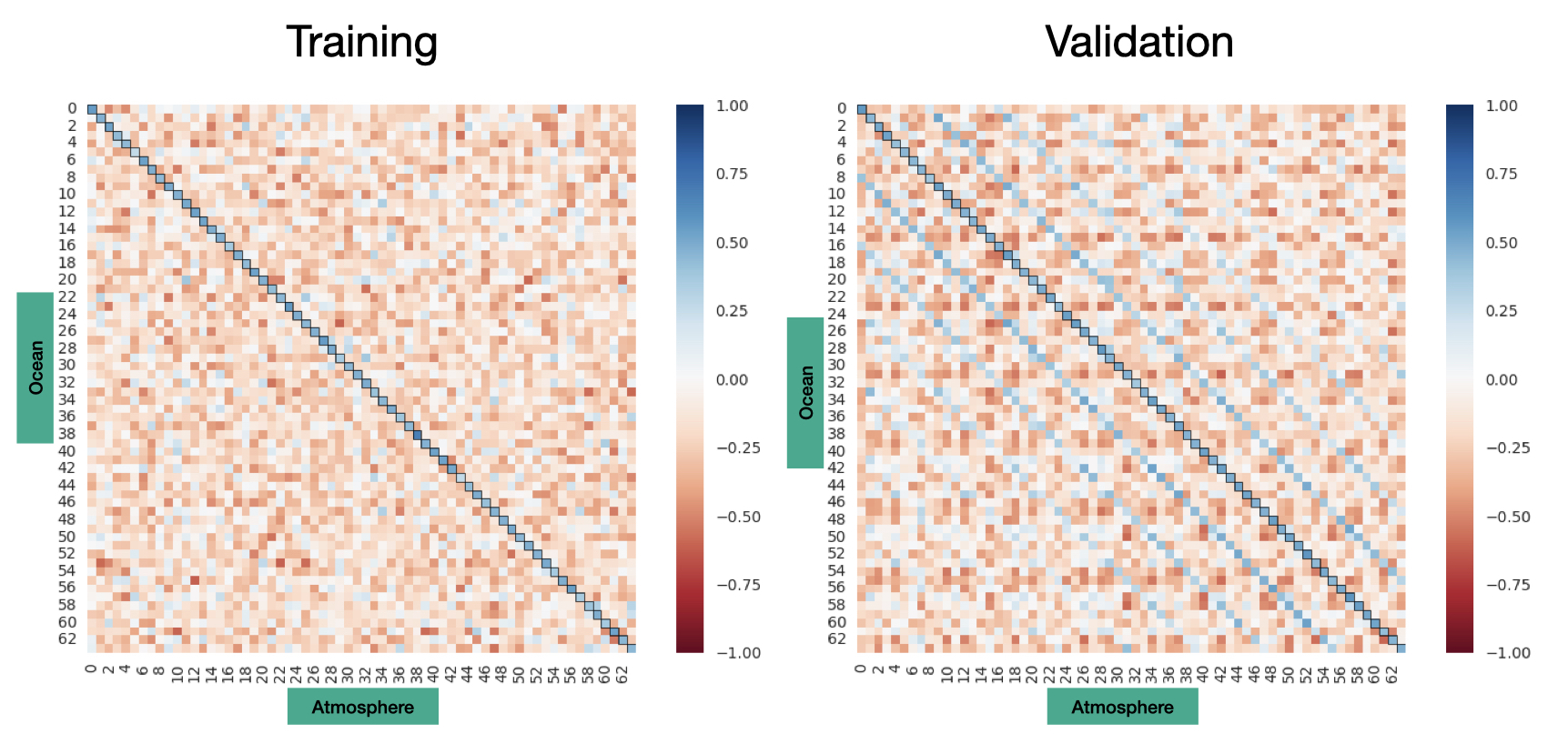}
        \caption{}
        \label{fig:cosine}
    \end{subfigure}
    \hfill
    \begin{subfigure}[t]{0.53\textwidth}
        \centering
        \includegraphics[width=\textwidth]{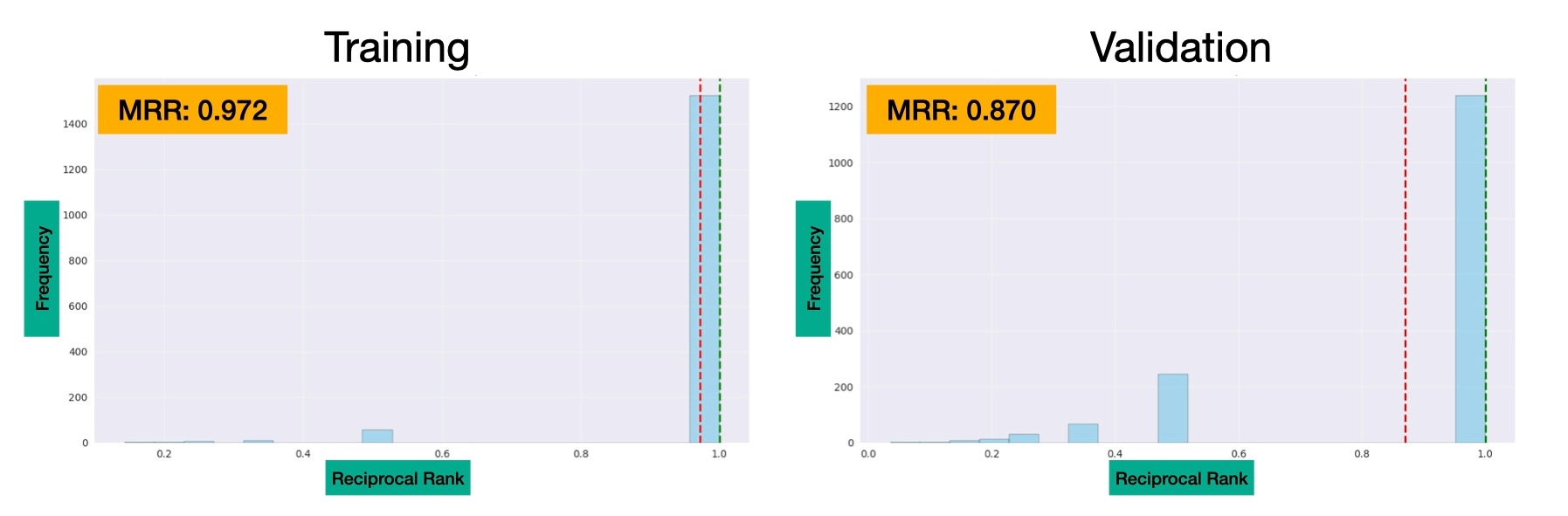}
        \caption{}
        \label{fig:mrr}
    \end{subfigure}
    \caption{\textbf{Pretraining metrics for \niva on training and validation data} (a) Cosine similarity matrix, where blue indicates high similarity and red indicates low similarity. The strong diagonal structure reflects correct alignment between paired ocean and atmospheric states in the latent space. (b) Reciprocal Rank (RR) distribution over 2000 samples. The concentration of samples at RR = 1 indicates that the model reliably retrieves the correct cross-modal pairs.}
\end{figure*}

\label{subsec-loss}

We use a variant of CLIP loss, replacing the standard cross-entropy loss with a Focal loss~\citep{lin2018focallossdenseobject} and applying it to ocean–atmosphere pairs. We assume a one-to-one correspondence between modalities within each temporal window and optimize the objective symmetrically in both directions (ocean→atmosphere and atmosphere→ocean). While this formulation enforces a single correspondence per sample, real-world ocean–atmosphere dynamics admit multiple plausible atmospheric realizations for a given ocean state. We discuss a generalized one-to-many extension in Sec.~\ref{appendix-loss}. To complement the contrastive objective, we incorporate a lightweight auxiliary reconstruction loss on the atmospheric modality, where the input field is reconstructed from its latent representation. This acts as a regularizer, encouraging retention of fine-grained spatial structure that may otherwise be suppressed under purely contrastive training. The reconstruction term is weighted by $\alpha \in [0, 0.1]$ to provide a stabilizing signal without dominating the primary objective. Together, these components yield representations that are globally aligned across modalities while preserving local spatial fidelity (see Sec.~\ref{appendix-equations-loss}).

\subsection{Post-Training}
\label{subsec-posttraining}
To assess whether our pretraining objective effectively learns latent spaces that capture cross-modal ocean-atmosphere dynamics, we evaluate the pretrained representations by decoding them with a simple linear head to predict a suite of major climate indices, including the Relative Oceanic Niño Index (RONI), Indian Ocean Dipole (IOD), Pacific-North American Pattern (PNA), Northern Annular Mode (NAM), North Atlantic Oscillation (NAO), and the Real-time Multivariate Madden-Julian Oscillation components (RMM1 and RMM2). These indices represent dominant modes of variability in the coupled Earth system and govern a wide range of medium-to long-term dynamical behavior, including teleconnections and large-scale oscillatory phenomena(see Sec.~\ref{appendix-derived-post} for more detail). 

This downstream evaluation tests whether the pretrained foundation model encodes physically meaningful and sufficiently expressive representations to reconstruct these indices from its learned joint latent space. Concretely, we freeze the ocean and atmospheric encoders and extract the ocean and atmosphere latent representations at each timestep, yielding two feature vectors of dimension $(B, 768)$, which are concatenated to form a joint representation of dimension $(B, 1536)$. This representation is then passed through a linear decoder to predict an output vector of dimension $(B, N)$, where $N$ denotes the number of climate indices considered. The task is a regression problem that is optimized using Huber loss~\citep{hastie2001elements}.


\label{sec-results}

\section{Results}
\begin{figure*}[t]
    \centering
    \includegraphics[width=\textwidth]{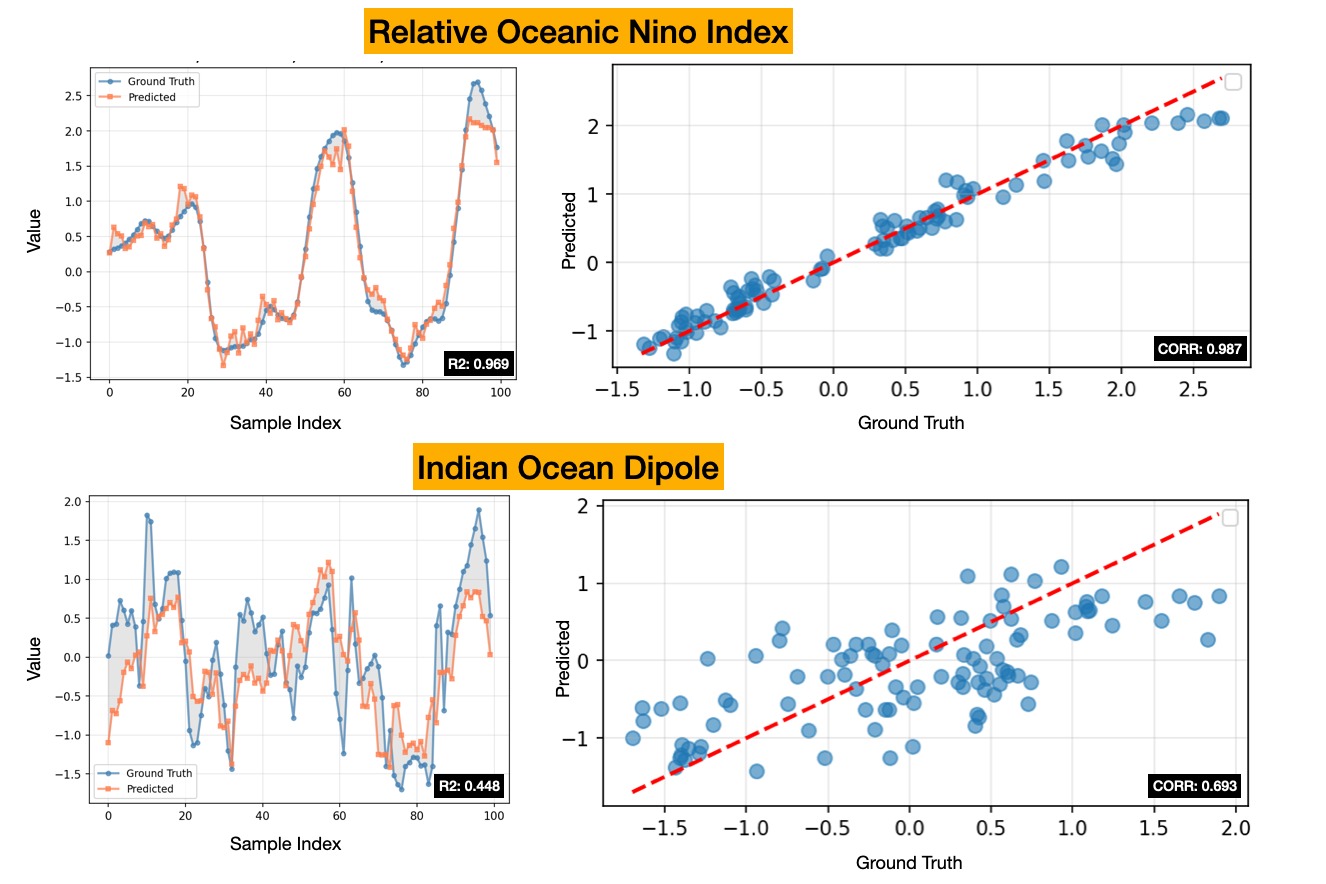}
\caption{\textbf{Post-training results for the RONI and IOD climate indices.} Line plots compare predicted values (orange) with ground-truth values (blue), while scatter plots show their correlations. RONI exhibits strong performance (R\textsuperscript{2} = 0.969), and IOD achieves moderate performance (R\textsuperscript{2} = 0.448).}
    
    \label{fig:roni}
\end{figure*}
\begin{figure*}[t]
\includegraphics[width=16cm]{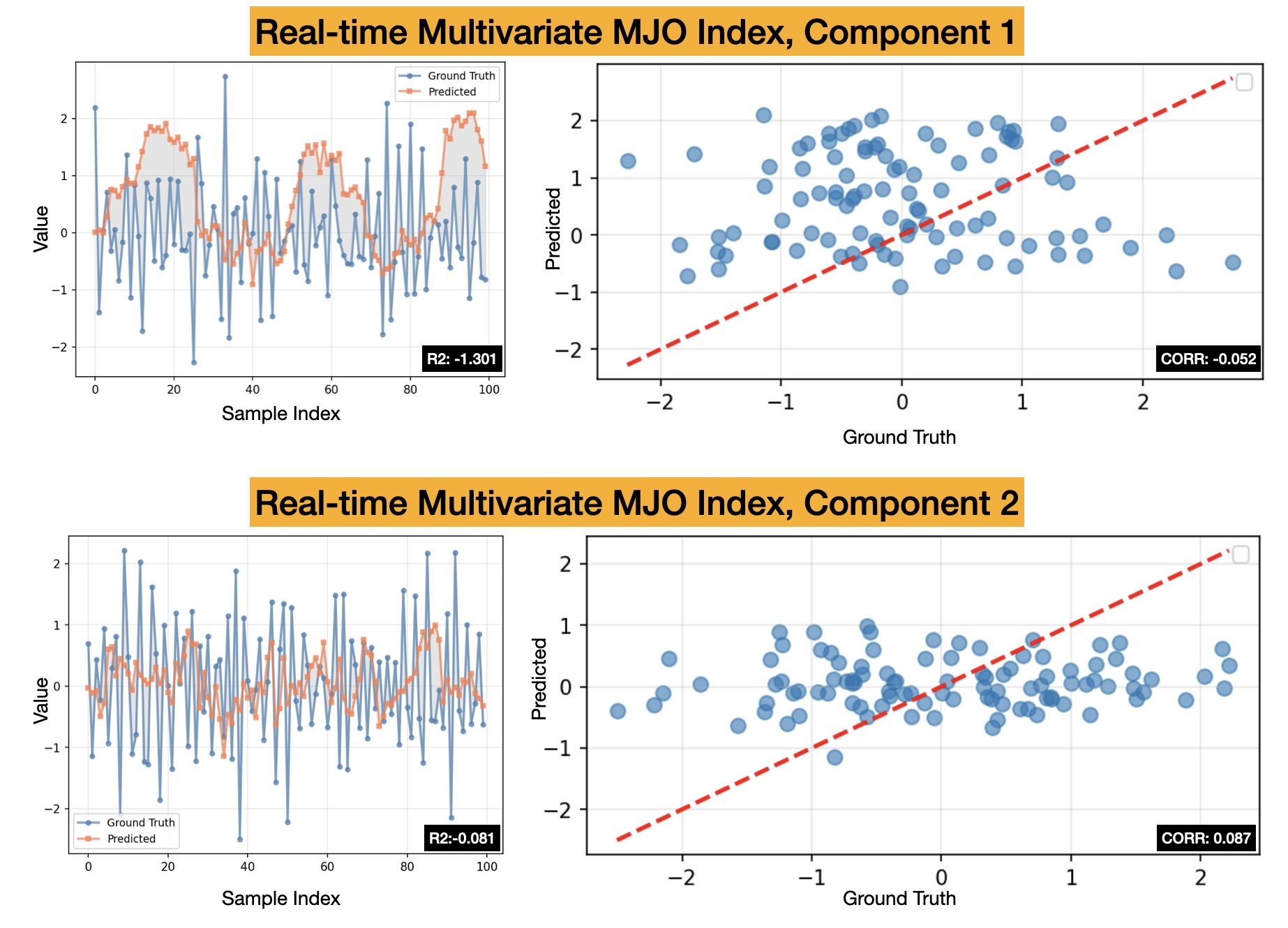}
\caption{\textbf{Post-training results for the Real-time multivariate MJO components 1 and 2 (RMM1 and RMM2).} Line plots compare predicted values (orange) with ground-truth values (blue), while scatter plots show their correlations. } 
\label{fig:mjo}
\end{figure*}
\subsection{Pre-training}
During pretraining, the model is optimized to align positive ocean–atmosphere pairs in a shared latent space while pushing apart negative pairs. Training is guided by our modified CLIP-style contrastive loss (Section 3.3.1). We use the Adam optimizer~\citep{kingma2017adammethodstochasticoptimization} with a learning rate of $0.001$ and train on $80$ NVIDIA A100 GPUs with a batch size of $4$ per GPU. Training is performed for up to 35 epochs with early stopping based on validation performance.

For this initial pretraining study, we restrict the temporal resolution to monthly aggregates. The training dataset consists of monthly samples from $1920–2000$ and $2040–2100$ for each of the $100$ ensemble members, while the period $2001–2015$ is held out as a test set, and $2016–2040$ is used for validation. This results in a total of $168,000$ training samples and $28,800$ validation samples. We track the loss as well as cosine similarity matrices and a ranking-based metric, Mean Reciprocal Rank (MRR), to gather interpretable measures of cross-modal alignment.

In the first evaluation, we visualize alignment using cosine similarity matrices computed over a fixed set of $64$ ocean–atmosphere pairs at each epoch for both training and validation splits (see Figure \ref{fig:cosine}). Ocean samples are arranged along the y-axis and atmospheric samples along the x-axis, with both modalities ordered identically to enable direct comparison along the diagonal. Blue denotes high similarity, while red denotes low similarity. In the best-performing model, the matrix exhibits a strong, well-defined blue diagonal, indicating that the model reliably aligns corresponding ocean and atmospheric states in the latent space. In addition, structured regions of elevated similarity appear off the diagonal, suggesting that a single ocean state may correspond to multiple dynamically consistent atmospheric realizations. This behavior reflects the inherently multi-modal nature of ocean–atmosphere coupling. Importantly, the dominance of the diagonal confirms correct pairwise alignment, while the presence of coherent off-diagonal patterns indicates that the model captures physically plausible alternatives rather than spurious correlations.

The second evaluation metric is the Mean Reciprocal Rank (MRR). For each ocean sample, the model ranks all candidate atmospheric states based on similarity in the learned latent space. If the correct atmospheric match appears in the  $k^{th}$ position, the reciprocal rank is $1/k$. This metric is well-suited to our setting, as it accounts for the fact that multiple atmospheric states may be physically consistent with a given ocean condition. Consequently, even when the exact temporal match is not ranked first, closely related atmospheric states may still occupy top positions. A high MRR therefore, indicates that the model consistently prioritizes the correct match along with other physically plausible candidates. In our experiments, we achieve a validation MRR of $0.91$ across all the validation samples, demonstrating strong alignment performance and confirming that the learned latent space effectively captures the underlying ocean–atmosphere relationships.

\subsection{Post-Training}
For post-training we train a simple linear decoder(see Sec.~\ref{subsec-posttraining}) on ERA5 data (see Sec.~\ref{subsec-data}. The linear decoder is trained on data from $1980-2000$ and tested on data from $2001-2015$ with $2016-2025$ as the validation set. Figure \ref{fig:roni} shows results for two representative indices from the post-training experiment, Relative Oceanic Niño Index (RONI) and Indian Ocean Dipole (IOD), achieving R\textsuperscript{2} scores of of $0.969$ and $0.448$ and a correlation of $0.98$ and $0.693$, respectively. 
RONI performs best among all the indices because it is primarily governed by oceanic variability and ocean–atmosphere coupling processes that our foundation model is designed to encode during pretraining (for additional discussion and results for all the indices, refer to Sec.~\ref{appendix-posttraining} in the Appendix).


\par
Figure \ref{fig:mjo} highlights two indices that perform most poorly in our post-training evaluation: the real-time multivariate Madden–Julian Oscillation (MJO) components, RMM1 and RMM2. This outcome can be attributed to two main factors. First, post-training is conducted at a monthly temporal resolution, which is too coarse to resolve MJO’s intraseasonal variability, typically operating on 30–60 day timescales. Second, MJO dynamics are not governed solely by oceanic initial conditions; they depend on a broader set of processes, including atmospheric humidity, convection, and large-scale circulation patterns. In some regions, land–atmosphere interactions such as soil moisture influences on convection can further modulate MJO propagation \citep{Zhang2010_FourMJOTheories}.
\par
These limitations highlight a clear opportunity for improvement. Incorporating additional Earth system components (e.g., land and sea ice), along with one-to-many temporal mapping that preserves higher-frequency atmospheric variability while maintaining lower-frequency ocean dynamics (see Sec.~\ref{appendix-pretraining}), will enable a more complete representation of processes governing the MJO and similar phenomena.




\label{sec-conclusion}
\section{Conclusion}
In this work, we introduced \niva, a foundation modeling framework for learning coupled Earth system dynamics. Focusing on the ocean–atmosphere subsystem, we learn a unified latent representation that captures large-scale climate variability. Across multiple variables and encoder architectures, our results show that NIVA learns consistent and physically meaningful cross-modal relationships, indicating that multimodal foundation models can indeed capture coupled dynamics across Earth system components.

Post-training evaluation further validates this approach. Fine-tuning the pretrained encoders to predict major climate indices yields strong performance, particularly on the Relative Oceanic Niño Index, demonstrating that joint ocean–atmosphere representation learning can recover key modes of climate variability. These results suggest that the learned latent space encodes the core dynamical structures governing medium- to long-range climate behavior and that pretraining on numerical simulations can transfer effectively to observational tasks.

At the same time, reduced performance on indices influenced by land and sea-ice processes highlights the limitations of the current two-modality setting. Addressing these gaps by incorporating additional Earth system components is a natural next step, enabling better cross-modal interactions and a physically complete foundation for climate applications.
\paragraph{Acknowledgements} This work was supported in part by the National Science Foundation under \AwardNumber. The project team gratefully acknowledges the contribution of Dr. Hansi Singh for her domain expertise and scientific guidance. We also acknowledge the Argonne Leadership Computing Facility (ALCF) at Argonne National Laboratory for providing computational resources that supported this work.

\bibliography{main}
\bibliographystyle{icml2026}

\newpage
\newpage
\appendix

\section{Data}

\subsection{Variables}
\label{appendix-variables}
\subsubsection{Pretraining}

\begin{table}[h]
\centering
\begin{tabular}{lp{0.6\linewidth}}
\toprule
\textbf{Type} & \textbf{Variable} \\
\midrule
Atmospheric & Sea-level pressure (slp) \\
Atmospheric & 2-m temperature (t2m) \\
Atmospheric & Zonal wind at 10 hPa (u10) \\
Atmospheric & Geopotential height at 500 hPa (z500) \\
Atmospheric & Eddy heat flux at 100 hPa (ehf100) \\
Atmospheric & Velocity potential at 200 hPa (vp200) \\
Atmospheric & Total column water vapor (tcwv) \\
Atmospheric & Outgoing longwave radiation (olr) \\
Atmospheric & Thickness between 300--700 hPa (z300m700) \\
Atmospheric & Stream function of surface wind stress (sf\_tau) \\
Atmospheric & Divergence of specific humidity at 850 hPa (div\_q850) \\
Ocean & Sea-surface temperature (sst) \\
Ocean & Relative Sea-surface temperature (rsst)\\
Ocean & Sea-surface height (ssh) \\
Ocean & Sea-surface salinity (sss) \\
Invariant & Surface geopotential (phis) \\
Invariant & Land fraction (landfrac) \\
Invariant & Insolation climatology (solin) \\
\bottomrule
\end{tabular}
\caption{Input variables used in NIVA pretraining.}
\end{table}

\subsubsection{Post-training}

\begin{table}[h]
\centering
\begin{tabular}{lp{0.6\linewidth}}
\toprule
\textbf{Type} & \textbf{Variable} \\
\midrule
Climate Index & Relative Oceanic Niño Index (RONI) \\
Climate Index & Indian Ocean Dipole (IOD) \\
Climate Index & Northern Annular Mode (NAM) \\
Climate Index & North Atlantic Oscillation (NAO) \\
Climate Index & Real-time Multivariate Madden-Julian Oscillation Index, component 1 (RMM1) \\
Climate Index & Real-time Multivariate Madden-Julian Oscillation Index, component 2 (RMM2) \\

\bottomrule
\end{tabular}
\caption{\niva post-training output variables}
\end{table}

\subsection{Preprocessing Steps}
\label{appendix-preprocessing}
\textbf{Data Ingestion.} Raw CESM2 and ERA5 outputs are standardized to a common structure. This includes consistent variable naming, longitude wrapping, removal of extraneous fields, conversion to float32, and chunking into multi-year segments to support efficient downstream processing.

\textbf{Computation of Derived Variables.} Several quantities not directly available in the model output are computed to enrich the feature space. These include wind-derived quantities (stream function, divergence, velocity potential), layer-thickness variables, moisture-related diagnostics, and climate indices. All derived fields follow the same spatial and temporal conventions as the native data, except for climate indices which are reduced spatially to a single dimension.

\textbf{Temporal Aggregation.} NIVA uses two aggregated products: (a) weekly aggregation of 52 seven-day windows each year. January 1 is removed to achieve 364 days per year, and (b) four-weekly aggregation, constructed by grouping the 52 weeks into 13 fixed four-week periods. These definitions are repeated annually and allow for consistent climatology for standardization.

\textbf{Spatial Restructuring.} Gridded variable data are restricted to a band of area covered by 60°S to 60°N. Data at high latitudes are removed for this phase of pretraining to avoid any influence of non-physical values associated with sea ice coverage. Furthermore, all gridded variables are regridded to a consistent 256 × 128 longitude–latitude grid.



\textbf{Standardization.} Anomaly fields are standardized relative to a 1940–2025 climatological baseline, chosen for compatibility with observation-based datasets (i.e., ERA5). Standardization can be weekly or four-weekly, depending on the aggregated product. Outputs include the standardized field, its climatological mean, and its standard deviation. Ocean datasets also include a consistent land–sea mask.

\subsection{Derived Variable Calculation}

\subsubsection{Pretraining}

\paragraph{Eddy heat flux at 100 hPa (ehf100)}
The eddy heat flux is the covariance of the meridional wind and temperature perturbations about their zonal mean \citep{andrews1987,newman2001}. At each grid point on the 100~hPa surface it is computed as
\begin{equation}
    \mathrm{ehf}_{100} = v'\,T',
    \qquad v' = v - [v], \qquad T' = T - [T],
\end{equation}
where $v$ (m~s$^{-1}$) is the meridional wind and $T$ (K) is the temperature at 100~hPa, and $[\,\cdot\,]$ denotes the zonal mean along a latitude circle.

\paragraph{Velocity potential at 200 hPa (vp200)}
The velocity potential is the scalar field whose Laplacian equals the horizontal wind divergence \citep{holton2013}. It is obtained on the 200~hPa surface by solving
\begin{equation}
    \nabla_{h}^{2}\,\mathrm{vp200} = \frac{\partial u_{200}}{\partial x} + \frac{\partial v_{200}}{\partial y},
\end{equation}
where $u_{200}$ and $v_{200}$ are the zonal and meridional wind components (m~s$^{-1}$) at 200~hPa, and $\nabla_{h}^{2}$ is the horizontal Laplacian on the sphere.

\paragraph{Thickness between 300--700 hPa (z300m700)}
Layer thickness is the difference in geopotential height between two pressure surfaces. It is computed directly as
\begin{equation}
    \mathrm{z300m700} = z_{300} - z_{700},
\end{equation}
where $z_{300}$ and $z_{700}$ are the geopotential heights (m) of the 300~hPa and 700~hPa surfaces, respectively.

\paragraph{Stream function of surface wind stress (sf\_tau)}
The stream function of the surface wind stress is the scalar field whose Laplacian equals the vertical component of the wind-stress curl \citep{holton2013}. It is obtained by solving
\begin{equation}
    \nabla_{h}^{2}\,\mathrm{sf\_tau} = \frac{\partial \tau_{y}}{\partial x} - \frac{\partial \tau_{x}}{\partial y},
\end{equation}
where $\tau_{x}$ and $\tau_{y}$ are the zonal and meridional components of the surface wind stress vector (N~m$^{-2}$).

\paragraph{Divergence of specific humidity at 850 hPa (div\_q850)}
The horizontal moisture flux divergence at 850~hPa quantifies low-level sources and sinks of water vapor and is computed from winds and specific humidity following \citet{wallace2006}:
\begin{equation}
    \mathrm{div\_q}_{850}
    = \frac{\partial (q_{850}\,u_{850})}{\partial x} + \frac{\partial (q_{850}\,v_{850})}{\partial y},
\end{equation}
where $q_{850}$ is the specific humidity (kg~kg$^{-1}$) and $u_{850}$ and $v_{850})$ are the zonal and meridional wind vectors (m~s$^{-1}$) at 850~hPa, respectively.

\subsubsection{Post-training}
\label{appendix-derived-post}
\paragraph{Relative Oceanic Nino Index (RONI)}
RONI is the operational ENSO index adopted by NOAA/CPC that corrects the traditional Oceanic Nino Index (ONI) for tropics-wide background warming by subtracting the tropical-mean SST anomaly, with a variance rescaling so that its amplitude is comparable to the ONI \citep{vanoldenborgh2021,lheureux2024}. Following the scaling-based formulation,
\begin{equation}
\begin{split}
    \mathrm{RONI} = {}& \bigl(\mathrm{SSTA}_{\mathrm{Nino3.4}} - \overline{\mathrm{SSTA}}_{\mathrm{trop}}\bigr) \\
    & \times \frac{\sigma_{\mathrm{Nino3.4}}}{\sigma_{(\mathrm{Nino3.4}-\overline{\mathrm{SSTA}}_{\mathrm{trop}})}}.
\end{split}
\end{equation}
where $\mathrm{SSTA}_{\mathrm{Nino3.4}}$ is the 3-month running mean SST anomaly averaged over the Nino-3.4 region ($5^{\circ}$S--$5^{\circ}$N, $170^{\circ}$W--$120^{\circ}$W), $\overline{\mathrm{SSTA}}_{\mathrm{trop}}$ is the 3-month running mean SST anomaly averaged over the global tropical belt ($20^{\circ}$S--$20^{\circ}$N), $\sigma_{\mathrm{Nino3.4}}$ is the standard deviation of the Nino-3.4 anomaly, and $\sigma_{\mathrm{Nino3.4}-\overline{\mathrm{SSTA}}_{\mathrm{trop}}}$ is the standard deviation of the unscaled difference. All anomalies are computed relative to the 1991--2020 climatology.

\paragraph{Indian Ocean Dipole (IOD)}
The IOD is characterized by the Dipole Mode Index \citep{saji1999}, defined as the difference in SST anomalies between the western and southeastern tropical Indian Ocean:
\begin{equation}
    \mathrm{DMI} = \overline{\mathrm{SSTA}}_{\mathrm{WTIO}} - \overline{\mathrm{SSTA}}_{\mathrm{SETIO}},
\end{equation}
where $\overline{\mathrm{SSTA}}_{\mathrm{WTIO}}$ is the area-averaged SST anomaly over the western tropical Indian Ocean ($10^{\circ}$S--$10^{\circ}$N, $50^{\circ}$E--$70^{\circ}$E) and $\overline{\mathrm{SSTA}}_{\mathrm{SETIO}}$ is the area-averaged SST anomaly over the southeastern tropical Indian Ocean ($10^{\circ}$S--$0^{\circ}$, $90^{\circ}$E--$110^{\circ}$E).

\paragraph{Northern Annular Mode (NAM)}
Following \citet{thompson1998}, the NAM index is defined as the leading principal component (PC) time series of monthly sea level pressure (SLP) anomalies poleward of $20^{\circ}$N. The leading empirical orthogonal function (EOF) is first computed from area-weighted SLP anomalies:
\begin{equation}
    \mathrm{SLP}'(\lambda,\varphi,t) = \sum_{k} \mathrm{PC}_{k}(t)\,\mathrm{EOF}_{k}(\lambda,\varphi),
\end{equation}
where $\mathrm{SLP}'$ is the SLP anomaly relative to the monthly climatology, $\lambda$ and $\varphi$ are longitude and latitude (with $\varphi \geq 20^{\circ}$N), and $t$ is time. The NAM index is then obtained by normalizing the leading PC by its standard deviation,
\begin{equation}
    \mathrm{NAM}(t) = \frac{\mathrm{PC}_{1}(t)}{\sigma_{\mathrm{PC}_{1}}}.
\end{equation}
Prior to the EOF calculation, each latitude is weighted by $\sqrt{\cos\varphi}$.

\paragraph{North Atlantic Oscillation (NAO)}
The NAO index is defined as the difference between normalized SLP anomalies at Lisbon, Portugal and Stykkish\'olmur/Reykjav\'ik, Iceland:
\begin{equation}
    \mathrm{NAO}(t) = \frac{\mathrm{SLP}'_{\mathrm{Lis}}(t)}{\sigma_{\mathrm{Lis}}} - \frac{\mathrm{SLP}'_{\mathrm{Ice}}(t)}{\sigma_{\mathrm{Ice}}},
\end{equation}
where $\mathrm{SLP}'_{\mathrm{Lis}}$ and $\mathrm{SLP}'_{\mathrm{Ice}}$ are the SLP anomalies at Lisbon and Iceland relative to the monthly climatology, and $\sigma_{\mathrm{Lis}}$ and $\sigma_{\mathrm{Ice}}$ are the corresponding long-term standard deviations of the station SLP anomalies used for normalization \citep{hurrell1995}.

\paragraph{Real-time Multivariate Madden--Julian Oscillation Index, component 1 (RMM1)}
RMM1 is the first principal component of the combined EOF analysis of near-equatorially averaged OLR, $850$~hPa zonal wind, and $200$~hPa zonal wind, following \citet{wheeler2004}. Daily fields are first averaged between $15^{\circ}$S and $15^{\circ}$N, the annual cycle and the most recent 120-day mean are subtracted, and each field is normalized by its global (longitude- and time-) variance $\sigma_{X}$. Projection onto the leading multivariate EOF then yields
\begin{multline}
    \mathrm{RMM1}(t) = \sum_{\lambda} \Bigl[
        \tfrac{\mathrm{OLR}'(\lambda,t)}{\sigma_{\mathrm{OLR}}}\,e^{\,\mathrm{OLR}}_{1}(\lambda) \\
      + \tfrac{u_{850}'(\lambda,t)}{\sigma_{u_{850}}}\,e^{\,u_{850}}_{1}(\lambda) \\
      + \tfrac{u_{200}'(\lambda,t)}{\sigma_{u_{200}}}\,e^{\,u_{200}}_{1}(\lambda)
    \Bigr].
\end{multline}
where $\lambda$ is longitude; $\mathrm{OLR}'$, $u_{850}'$, and $u_{200}'$ are the $15^{\circ}$S--$15^{\circ}$N-averaged filtered anomalies (W~m$^{-2}$ and m~s$^{-1}$, respectively); $e^{X}_{1}(\lambda)$ is the longitudinal structure of EOF~1 for field $X$; and $\sigma_{X}$ is the corresponding normalization factor. The result is divided by the standard deviation of the 1979--2001 time series so that RMM1 has unit variance.

\paragraph{Real-time Multivariate Madden--Julian Oscillation Index, component 2 (RMM2)}
RMM2 is constructed identically to RMM1 but projects the daily filtered anomalies onto the second multivariate EOF, which is in approximate quadrature with EOF~1 \citep{wheeler2004}:
\begin{multline}
    \mathrm{RMM2}(t) = \sum_{\lambda} \Bigl[
        \tfrac{\mathrm{OLR}'(\lambda,t)}{\sigma_{\mathrm{OLR}}}\,e^{\,\mathrm{OLR}}_{2}(\lambda) \\
      + \tfrac{u_{850}'(\lambda,t)}{\sigma_{u_{850}}}\,e^{\,u_{850}}_{2}(\lambda) \\
      + \tfrac{u_{200}'(\lambda,t)}{\sigma_{u_{200}}}\,e^{\,u_{200}}_{2}(\lambda)
    \Bigr],
\end{multline}
where $e^{X}_{2}(\lambda)$ is the longitudinal structure of EOF~2 for each input field, and all other quantities are as defined for RMM1. The pair (RMM1, RMM2) spans the two-dimensional MJO phase space.

\section{\niva}
\subsection{Pretraining}
\label{appendix-pretraining}
To account for different temporal resolutions for atmospheric and oceanic states we also include the theory behind One-to-many temporal mapping which will be used for future experiments for richer latent representations. The method described in ~\ref{subsec-pretraining} is One-to-one temporal mapping.
\paragraph{One-to-many temporal mapping.}
To account for the differing intrinsic timescales of the two systems, we also consider a setting where atmospheric data are available at higher temporal resolution (e.g., weekly) compared to ocean data (e.g., monthly). In this case, multiple atmospheric states correspond to a single ocean state within a shared temporal window. Concretely, the atmospheric input is structured as $(k, C, 128, 256)$ per sample (e.g., $k=4$ weeks per month). During training, this tensor is reshaped to $(\text{batch size} \times k, C, 128, 256)$ and passed through the atmospheric encoder, producing latent representations of shape $(\text{batch size} \times k, d)$. These are then reshaped back to $(\text{batch size}, k, d)$ and  jointly compared against the corresponding ocean latent representations of shape $(\text{batch size}, d)$.
\begin{table}[h]
\centering
\begin{tabular}{l l c}
\hline
\textbf{Ocean Encoder} & \textbf{Atmos. Encoder} & \textbf{Memory (MiB)} \\
\hline
SFNO & SFNO  & 10795 \\
ViT  & ViT   & 5650  \\
ViT  & SwinV2 & 4063  \\
ViT  & SFNO   & 8241  \\
ViT  & CaFA   & 7301  \\
\hline
\end{tabular}
\caption{Pretraining memory usage across encoder configurations. Experiments conducted on a Tesla T4 GPU with 2 CPU cores, 7.3 GB RAM, and batch size of 2. Here Atmos. refers to Atmospheric. }
\label{tab:training_specs}
\vspace{-3em}
\end{table}
\subsection{Ocean \& Atmospheric Encoders}
\label{appendix-encoder}
\paragraph{Transformer-based architectures.}
For the ViT-based ocean encoder, we adopt a masking strategy inspired by MAE-ViT~\citep{He2022_MAE}, applied at the tokenization stage. Specifically, we construct a fixed land–ocean mask derived from land-surface fraction. Tokens with an aggregated land fraction exceeding 1\% are discarded, ensuring that the encoder operates exclusively on spatial regions containing valid ocean observations. To further reduce boundary artifacts, we address partial land contamination within retained tokens by extrapolating ocean variables across coastal regions, smoothing transitions and preventing sharp discontinuities at land–sea interfaces. For the Swin-based ocean encoder, we introduce a prior-masked hierarchical encoding strategy that leverages Swin’s window-based attention mechanism under the pre-computed land-ocean mask prior which is used to determine token validity. Windows with more than 90\% land coverage are treated as invalid and excluded from attention computation, inducing a structured sparsity pattern over the spatial token grid. For the remaining windows, self-attention is computed only over valid ocean tokens (land fraction $< 1\%$) following Swin’s local window attention design, while invalid land tokens are excluded from computation but retained implicitly through the fixed grid structure to preserve spatial alignment across windows. This formulation preserves Swin’s hierarchical and shifted-window information flow while introducing a sparsity prior that focuses representation learning on ocean-dominated and coastal transition regions.

Across architectures, SFNO demonstrates the strongest representational performance for both oceanic and atmospheric variables, consistent with its ability to capture global spatial dependencies via spectral operators. However, this comes at a higher computational cost and requires additional adaptations to handle sparsity. In contrast, masked transformer-based approaches provide a more computationally efficient alternative, reducing the effective compute associated with ocean inputs by approximately 30\% while using lighter architectures (see Table~\ref{tab:training_specs}).
\subsubsection{Multi Target Loss}
\label{appendix-loss}
To account for multiple positive match pairs during pretraining we propose two different loss formulations building on the contrastive loss discussed in Sec.~\ref{subsec-loss}
\paragraph{Different Temporal Resolutions}
Given the inherently more chaotic nature of atmosphere compared to ocean, it is prudent to assume a formulation  where we consider a lower temporal frequency(e.g weekly) for ocean (e.g monthly) compared to atmosphere to reduce loss of information (see Sec.~\ref{appendix-pretraining}). Under such conditions, a single ocean state corresponds to multiple valid atmospheric states within the same aggregation window. To account for this, we replace the standard one-hot target with a multi-hot target distribution over all valid atmospheric indices. Specifically, for each ocean state $i$, the corresponding target vector contains $k$ positive entries, where
\begin{equation}
k = \frac{\text{temporal resolution of atmospheric states}}{\text{temporal resolution of ocean states}}.
\end{equation}

For example, in the case of monthly ocean states and weekly atmospheric states, each ocean state is associated with $k = 4$ atmospheric states. The target distribution assigns equal weight to these valid matches, while all other entries are treated as negatives. Accordingly, the cross-entropy loss includes $k$ positive terms instead of a single positive term and is normalized by $k$ to ensure consistent scaling across different temporal aggregation settings. 

\paragraph{Same Temporal Resolution via Soft Targets}

For this formulation, we have the ocean and atmospheric states at the same temporal resolution(e.g., monthly). We assume that while an ocean and atmospheric state at the same timestep form a valid match, multiple other atmospheric states may also correspond to the same ocean state. To achieve this, we construct soft target distributions using the Relative Oceanic Niño Index (RONI) as a measure of similarity between ocean-atmosphere pairs. Let $\Delta_{\text{RONI}} \in [0,1]$ denote the normalized difference between the RONI values of an ocean-atmosphere pair. We define the target similarity as
\begin{equation}
s_{ij} = 1 - \Delta_{\text{RONI}}(i,j),
\end{equation}
where higher values indicate stronger correspondence. We then threshold this similarity to distinguish valid matches from negatives. Specifically, if
\begin{equation}
s_{ij} < 0.5,
\end{equation}
the pair is treated as a true negative and assigned zero weight. Otherwise, the pair is considered a valid match. Consequently, instead of a binary target matrix of ones and zeros, we construct a soft target distribution in which the strength of correspondence between two states is determined by $s_{ij}$.

We then apply a cross-entropy loss over these soft targets, normalized by the total target weight to ensure consistent scaling. This formulation enables the model to account for multiple physically plausible matches, assigning higher weights to strongly aligned ocean--atmosphere pairs while suppressing weak or spurious correspondences. This formulation was not used to test the post-training objective of predicting RONI to ensure there was no data leakage and spurious results. 
\section{Experiments}
\begin{figure*}[h]
\includegraphics[width=16cm]{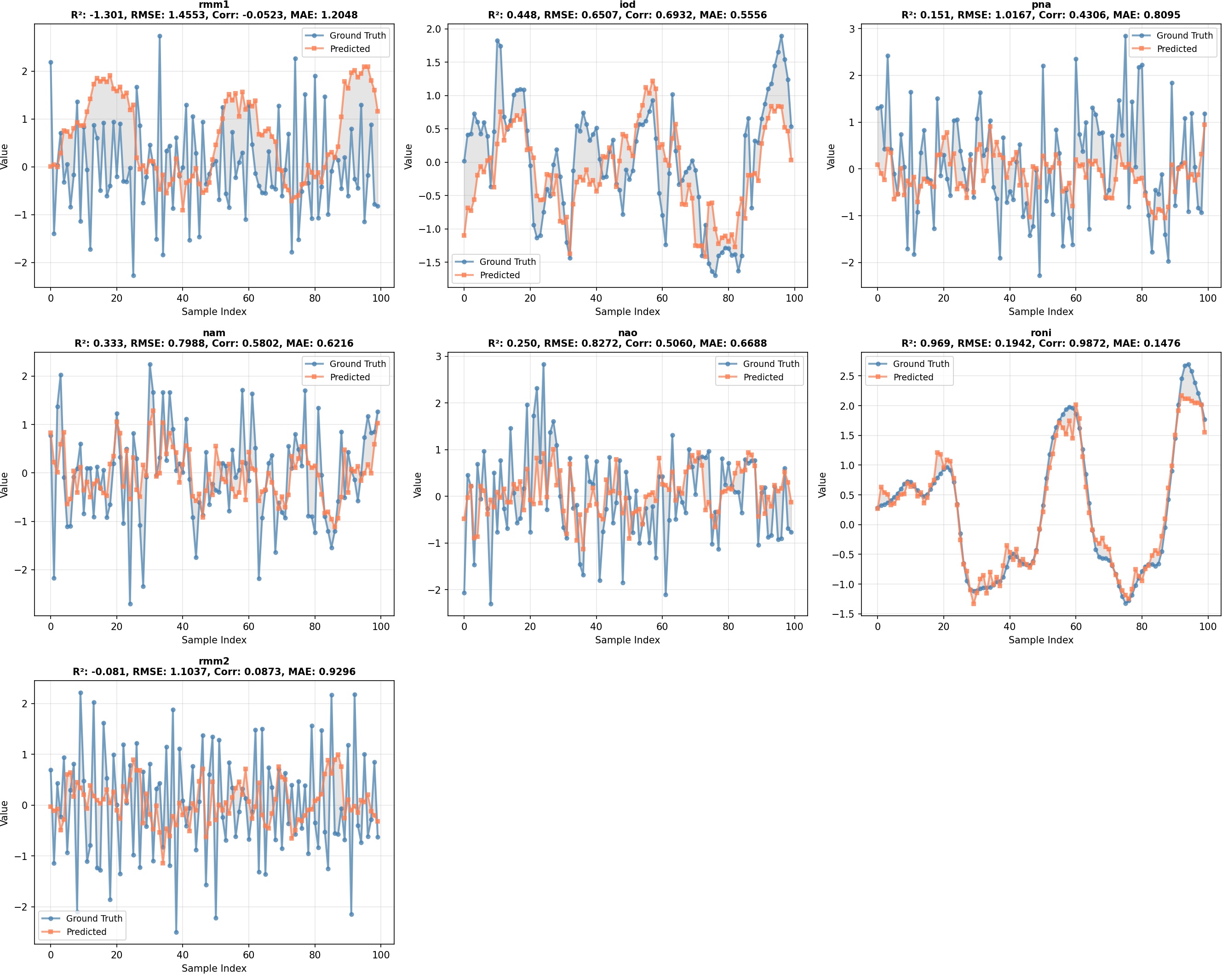}
\caption{Post-training results for all the climate indices. Line plots compare predicted values (orange) with ground-truth values (blue).} 
\label{fig:appendix1}
\end{figure*}
\begin{figure*}[h]
\includegraphics[width=16cm]{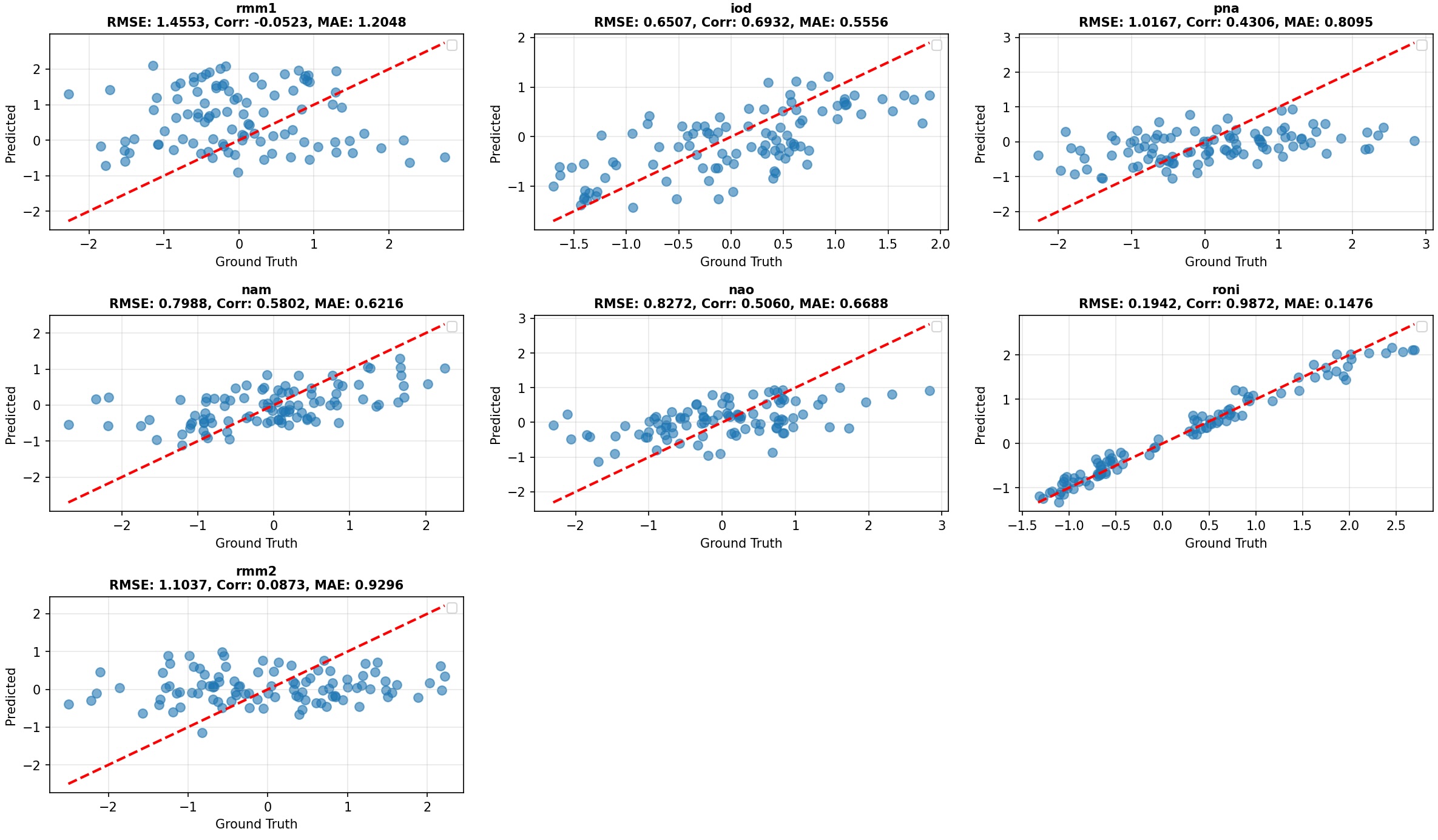}
\caption{Post-training results for all the climate indices. Scatter plots show the correlations between ground truth and predicted values.} 
\label{fig:appendix2}
\end{figure*}
\subsection{Pre-training}
\subsection{Post-Training}
\label{appendix-posttraining}
In contrast to RONI, several of the other indices include additional sources of variability originating from land, sea-ice processes, and higher-frequency atmospheric noise, making them difficult to construct from monthly aggregated ocean–atmosphere features alone. However, even for these more challenging indices (specify indices), the model demonstrates a reasonable ability to capture the positive and negative phases of the variability signal (Fig. \ref{fig:appendix1}). From a scientific perspective, this is often more important than accurately reproducing the exact magnitude of the index, as many downstream climate impacts and teleconnections depend primarily on the sign and relative amplitude of anomalies rather than their absolute values. The model’s ability to recover these phases suggests that the latent space encodes physically meaningful structure.

\section{Equations}
\subsection{Loss}
\label{appendix-equations-loss}
\begin{align}
\mathcal{L}_{\text{total}} 
= \mathcal{L}_{\text{FL}}^{o \rightarrow a} + \mathcal{L}_{\text{FL}}^{a \rightarrow o}
+ \alpha \, \mathcal{L}_{\text{recon}}.
\end{align}
\noindent where $B$ denotes the batch size. Let $\mathbf{o}_i \in \mathbb{R}^{d}$ and $\mathbf{a}_i \in \mathbb{R}^{d}$ denote the ocean and atmospheric embeddings for the $i$-th sample, where $d$ is the latent embedding dimension.

We define the similarity (logit) matrix $\mathbf{Z} \in \mathbb{R}^{B \times B}$ as
\[
Z_{ij} = \frac{\mathrm{cos}(\mathbf{o}_i, \mathbf{a}_j)}{\tau},
\]
where $\mathrm{cos}(\cdot,\cdot)$ denotes cosine similarity and $\tau$ is a temperature parameter.

The prediction distribution is obtained via row-wise softmax:
\[
P_{ij} = \frac{\exp(Z_{ij})}{\sum_{k=1}^{B} \exp(Z_{ik})}.
\]

The target matrix $\mathbf{Y} \in \mathbb{R}^{B \times B}$ is defined as a one-hot identity matrix:
\[
Y_{ij} =
\begin{cases}
1, & \text{if } i = j, \\
0, & \text{otherwise}.
\end{cases}
\]

We adopt a Focal Loss formulation to address class imbalance and emphasize hard negatives. The ocean-to-atmosphere loss is defined as
\[
\mathcal{L}_{\text{FL}}^{o \rightarrow a}
= - \frac{1}{B} \sum_{i=1}^{B} \sum_{j=1}^{B}
(1 - P_{ij})^{\gamma} \, Y_{ij} \log P_{ij},
\]
where $\gamma \geq 0$ is the focusing parameter.

The atmosphere-to-ocean loss $\mathcal{L}_{\text{FL}}^{a \rightarrow o}$ is defined analogously by swapping $\mathbf{o}$ and $\mathbf{a}$.

Finally, $\mathcal{L}_{\text{recon}}$ denotes the reconstruction loss over the atmospheric modality. Let $\mathbf{x}^a_i \in \mathbb{R}^{C \times H \times W}$ be the ground-truth atmospheric field and $\hat{\mathbf{x}}^a_i \in \mathbb{R}^{C \times H \times W}$ its reconstruction from the latent representation. Then,
\[
\mathcal{L}_{\text{recon}} = \frac{1}{B} \sum_{i=1}^{B} \left\| \hat{\mathbf{x}}^a_i - \mathbf{x}^a_i \right\|_2^2,
\]
where $C$ denotes the number of atmospheric channels and $H \times W$ is the spatial resolution.

\onecolumn

\end{document}